\documentclass[letterpaper, 10 pt, conference]{ieeeconf}  

\IEEEoverridecommandlockouts                              

\usepackage{graphicx}    
\usepackage{epsfig} 
\usepackage{times} 
\usepackage{amsmath} 
\usepackage{amssymb}  
\usepackage{threeparttable}
\usepackage{booktabs}

\usepackage{footnote} 
\usepackage[]{siunitx}
\DeclareSIUnit{\degree}{°}
\DeclareSIUnit{\deg}{deg}
\DeclareSIUnit{\nothing}{\relax}
\DeclareSIUnit{\op}{Op}
\DeclareSIUnit{\pixel}{px}
\DeclareSIUnit{\percent}{\%}
\DeclareSIUnit{\fps}{frame/s}

\usepackage{color}
\usepackage{tabularx}    
\usepackage{tabularray}
\usepackage{array}
\usepackage{booktabs}
\usepackage{multirow}
\usepackage{subcaption}

\usepackage{utfsym}

\makeatletter
\newcommand{\todo}[1]{\textit{\color{red}\textbf{TODO}~#1}\@latex@warning{TODO: #1}}
\makeatother

\usepackage{hyperref}

\definecolor{somegray}{rgb}{0.5, 0.5, 0.5}
\newcommand{\darkgrayed}[1]{\textcolor{somegray}{#1}}
\makeatletter
\newcommand*\titleheader[1]{\gdef\@titleheader{#1}}
\AtBeginDocument{%
  \let\st@red@title\@title
  \def\@title{%
    \vskip-3.0em
    \bgroup\normalfont\large\centering\@titleheader\par\egroup
    \vskip0.0em\st@red@title}
}
 
\makeatother
 
\titleheader{\darkgrayed{This paper has been accepted for publication in the IEEE ICRA 2026 conference\\\copyright 2026 IEEE.}}

\title{\LARGE \bf
Tiny-DroNeRF: Tiny Neural Radiance Fields aboard Federated Learning-enabled Nano-drones
}

\author{Ilenia Carboni$^{1}$$^{3}$$^{*}$, Elia Cereda$^{1}$$^{*}$, Lorenzo Lamberti$^{1}$$^{2}$$^{*}$, Daniele Malpetti$^{1}$, Francesco Conti$^{3}$, Daniele Palossi$^{1}$$^{2}$
\thanks{This work was partially supported by the SNSF RoboMix2 project (grant nb. 10004854).}
\thanks{$^{*}$ denotes equal contributions.}%
\thanks{$^{1}$I. Carboni, E. Cereda, L. Lamberti, D. Malpetti and D. Palossi are with the Dalle Molle Institute for Artificial Intelligence, USI-SUPSI, Switzerland. Corresponding author: {\tt\small ilenia.carboni@supsi.ch}.}%
\thanks{$^{2}$ L. Lamberti and D. Palossi are also with the Integrated Systems Laboratory, ETH Zürich, Switzerland.}%
\thanks{$^{3}$ I. Carboni and F. Conti are with the Department of Electrical, Electronic and Information Engineering, University of Bologna, Italy.}%
}

\begin{document}

\maketitle
\thispagestyle{empty}
\pagestyle{empty}

\begin{abstract}
Sub-\SI{30}{\gram} nano-sized aerial robots can leverage their agility and form factor to autonomously explore cluttered and narrow environments, like in industrial inspection and search and rescue missions.
However, the price for their tiny size is a strong limit in their resources, i.e., sub-\SI{100}{\milli\watt} microcontroller units (MCUs) delivering $\sim$\SI{100}{\giga Ops/\second} at best, and memory budgets well below \SI{100}{\mega\byte}.
Despite these strict constraints, we aim to enable complex vision-based tasks aboard nano-drones, such as dense 3D scene reconstruction: a key robotic task underlying fundamental capabilities like spatial awareness and motion planning.
Top-performing 3D reconstruction methods leverage neural radiance fields (NeRF) models, which require \SI{}{\giga\byte}s of memory and massive computation, usually delivered by high-end GPUs consuming 100s of Watts.
Our work introduces Tiny-DroNeRF, a lightweight NeRF model, based on Instant-NGP, and optimized for running on a GAP9 ultra-low-power (ULP) MCU aboard our nano-drones.
Then, we further empower our Tiny-DroNeRF by leveraging a collaborative federated learning scheme, which distributes the model training among multiple nano-drones.
Our experimental results show a 96\% reduction in Tiny-DroNeRF's memory footprint compared to Instant-NGP, with only a \SI{5.7}{\decibel} drop in reconstruction accuracy.
Finally, our federated learning scheme allows Tiny-DroNeRF to train with an amount of data otherwise impossible to keep in a single drone's memory, increasing the overall reconstruction accuracy.
Ultimately, our work combines, for the first time, NeRF training on an ULP MCU with federated learning on nano-drones.
\end{abstract}

\section*{Supplementary video material}
Supplementary video can be found at: \\
\url{https://youtu.be/-frFPUBGa0c}


\section{Introduction} \label{sec:introduction}

\begin{table*}[t]
\centering
\caption{Comparison of novel view synthesis methods on the Synthetic NeRF 360° dataset~\cite{mildenhall2020nerf}.}
\label{tab:nerf-background}

\resizebox{\linewidth}{!}{
\renewcommand{\arraystretch}{1.15}
\setlength{\tabcolsep}{1.7pt}
\setlength{\tabcolsep}{3pt}
\begin{tabular}{l c c l c c c c c c c}
\toprule
\textbf{Method} & \textbf{Year} & \textbf{Neural} & \textbf{Key contributions} &
\textbf{\begin{tabular}[c]{@{}c@{}}PSNR \\ {[dB]}\end{tabular}} &
\textbf{\begin{tabular}[c]{@{}c@{}}SSIM \\ {[-]}\end{tabular}} &
\textbf{\begin{tabular}[c]{@{}c@{}}Params \\ {[\SI{}{\mega\byte}]}\end{tabular}} &
\textbf{\begin{tabular}[c]{@{}c@{}}Training \\ time\end{tabular}} &
\textbf{\begin{tabular}[c]{@{}c@{}}Power \\ class {[\SI{}{\watt}]}\end{tabular}} &
\textbf{Device} \\
\midrule

\textbf{NeRF}~\cite{mildenhall2020nerf} & 2021 & \checkmark & Coordinate-based MLP & 31.01 & 0.947 & 5 & 24 h &
$10^2$ & GPU \\

\textbf{SNeRG}~\cite{hedman2021snerg} & 2021 & \checkmark & Neural grid & 30.38 & 0.950 & 1772 & 15 h & 
$10^2$ & GPU \\

\textbf{PlenOctrees}~\cite{plenoctrees2021} & 2021 & \checkmark & Octree + Spherical harmonics & 31.71 & 0.958 & 1930 & 15 h &
$10^2$ & GPU \\

\textbf{Plenoxels}~\cite{plenoxels2022} & 2022 & $\times$ & Voxel grid + neural network-free & 31.71 & 0.958 & 778 & 11 min & 
$10^2$ & GPU \\

\textbf{Instant-NGP}~\cite{mueller2022instant} & 2022 & \checkmark & Parametric hash-grid  & \textbf{33.18} & 0.959 & 46 & 5 min & 
$10^2$ & GPU \\

\textbf{TensoRF}~\cite{tensorf2022} & 2022 & $\times$ & Voxel grid w/ tensor factorization & 33.14 & \textbf{0.963} & 72 & 17 min &
$10^2$ & GPU \\

\textbf{K-Planes}~\cite{fridovich2023k} & 2023 & $\times$ & Multi-planes & 32.21 & 0.960 & 385 & 38 min &
$10^2$ & GPU \\

\midrule

\textbf{MobileNeRF}~\cite{mobilenerf2023} & 2023 & \checkmark & Rasterized mesh (baked) & 30.90 & 0.947 &
538 &
\begin{tabular}[c]{@{}l@{}}27 h\end{tabular} &
$10^2$ & GPU \\

\textbf{BiRF}~\cite{birf2023} & 2023 & \checkmark & 3D-2D binarized feature hash encoding grid & 32.03 & 0.954 & 0.5 &
$>5$ min & $10^2$ & GPU \\ 

\textbf{CNC}~\cite{cncnerf2024} & 2024 & \checkmark & Context-based compression framework & 33.19 & 0.964 & 0.418 & $>5$ min
& $10^2$ & GPU \\

\midrule

\textbf{Tiny-DroNeRF} & 2025 & \checkmark & Instant-NGP-based, tailored for MCUs & 26.94 & 0.892 & \textbf{0.284} & \textbf{28s} & $10^2$ & GPU \\
(ours) & & & & 26.94 & 0.892 & \textbf{0.284} & 97 min & $\mathbf{<0.10}$ & \textbf{MCU} \\

\bottomrule
\end{tabular}
}
\end{table*}

Autonomous flying miniaturized nano-drones, with a few tens of grams of weight and sub-\SI{10}{\centi\meter} diameter, can cope with cluttered and narrow spaces, such as hazardous industrial plants~\cite{uav_chemicals} or post-disaster environments~\cite{UAV_safety_rescue}.
Thanks to their autonomy, i.e., all processing is carried onboard with no external infrastructure, they are not limited to line-of-sight operations or to a maximum distance from the base station.
Nano-drones equipped with small cameras and onboard intelligence have been demonstrated in simple inspection and detection tasks, including anomaly detection~\cite{zauliExploitingNanoAerial2024}, semantic localization~\cite{zimmermanFullyOnboardLowPower2024}, and object detection~\cite{lamberti_bio_inspired_2023}.
Their effectiveness further increases in swarm configurations, where multiple drones cover larger areas simultaneously or collaborate toward a shared goal~\cite{pourjabarMultisensoryAnticollisionDesign2023}, such as survivor search.
However, more complex autonomous workloads are still infeasible on this class of tiny robots.
A primary reason is their lack of spatial awareness, stemming from the limited computational and sensory resources they can carry.
Compared to \SI{}{\kilo\gram}-scale flying drones, a nano-drone can rely on roughly 1000$\times$ less memory and computation~\cite{lamberti2023simtoreal}, and simpler sensors, such as low-resolution monochrome monocular cameras.

Neural radiance fields (NeRF)~\cite{mildenhall2020nerf} are a well-established paradigm for dense 3D reconstruction tasks.
Enabling NeRF on nano-drones would unlock the ability to autonomously perform novel view synthesis~\cite{mildenhall2020nerf}, SLAM~\cite{coslam}, 3D surface reconstruction~\cite{liNeuralangeloHighFidelityNeural2023}, or 3D semantic scene reconstruction~\cite{roldao3DSemanticScene2022}.
However, current NeRF algorithms demand memory and compute resources that exceed nano-drones' capabilities by orders of magnitude~\cite{mueller2022instant}, due to their complexity and the need for scene-specific retraining.
Therefore, they are generally limited to offline execution on GPUs, with \si{\giga\byte}s of memory available, and consuming hundreds of Watts.

In this work, we address the task of novel view synthesis using NeRF under the constraints of the limited resources of an ultra-low-power MCU suitable for miniaturized nano-drones.
Taking advantage of the tiny size of nano-drones, we focus on the operation of a swarm, where each nano-drone takes a few images of the target object from different known poses and trains a separate NeRF instance onboard.
Then, thanks to a federated learning approach~\cite{mcmahan2017communication}, we iteratively aggregate the individual NeRF instances to create a global NeRF producing a richer 3D reconstruction than any single drone can provide.
Overall, our work enables the cooperative construction of a continuous 3D representation of the scene, enabling the generation of photorealistic views from previously unseen camera poses using images captured from non-overlapping drone viewpoints. 

In detail, our paper's contributions are:
\begin{itemize}
    \item a lightweight NeRF algorithm, called \textit{Tiny-DroNeRF}, based on the State-of-the-Art (SoA) Instant-NGP~\cite{mueller2022instant}. Through hyperparameter optimization, we seek the best trade-off between 3D reconstruction quality, i.e., peak signal to noise ratio (PSNR), memory, and computation.
    \item A memory-efficient implementation of Tiny-DroNeRF, targeting the ultra-low-power multi-core GreenWaves Technologies (GWT) GAP9 MCU aboard a nano-drone.
    \item A detailed experimental analysis of \textit{i}) the reconstruction accuracy on standard synthetic datasets~\cite{mildenhall2020nerf} and on a novel real-world nano-drone dataset, and \textit{ii}) the execution time and memory usage on the GAP9 MCU.
    \item A federated learning scheme for distributed training of Tiny-DroNeRF across a swarm of nano-drones.
\end{itemize}

In our experiments, Tiny-DroNeRF drastically cuts resource requirements while maintaining useful reconstruction quality.
Compared to Instant-NGP~\cite{mueller2022instant}, we reduce memory usage by 96\% (from \SI{527}{\mega\byte} to \SI{21.4}{\mega\byte}), while losing only \SI{5.7}{\decibel} of PSNR.
We complete a full forward-backward step in \SI{73}{\milli\second}, i.e., \SI{97}{\minute} for a complete training on a single nano-drone (\SI{10}{\kilo\nothing} steps).
Through our efficient hardware-aware algorithm, we succeed in making the workload compute-bound.
Our implementation exhibits excellent memory locality, requiring only \SI{3.6}{\mega\byte} of data transfers to/from off-chip memory per computational step.
In field tests, we reconstruct a real-world scene from the nano-drone's onboard grayscale images at up to 21.2 PSNR.

Moreover, we demonstrate that our federated learning scheme allows each nano-drone, which carries only a limited set of local images, to synthesize unseen portions of the scene by leveraging the knowledge of its peers, without exchanging raw training images.
This federated approach achieves a PSNR just \SI{0.6}{\decibel} below a centralized training with access to all images, while requiring only \SI{2.24}{\second} of communication overhead per merge (done every \SI{1}{\kilo\nothing} local training steps).
To the best of our knowledge, this is the first demonstration of on-device NeRF tailored to highly constrained computing devices and the first application of a federated learning scheme on a swarm of resource-constrained nano-drones.
\section{Related work} \label{sec:related_work}

\subsection{Novel view synthesis methods}
Neural radiance fields~\cite{mildenhall2020nerf} introduced a paradigm for representing 3D scenes as continuous functions, enabling photorealistic novel-view synthesis and high-fidelity scene reconstruction. This approach established implicit neural representations, in which a multi-layer perceptron (MLP) encodes a continuous volumetric scene that can be queried at arbitrary 3D coordinates, and made this a widely adopted technique. Nevertheless, vanilla NeRF suffers from slow training and rendering, motivating extensive research aimed at accelerating both optimization and inference.

To address these limitations, subsequent work has explored explicit scene parameterizations as a means of improving efficiency. 
For example, PlenOctrees~\cite{plenoctrees2021} precompute a sparse octree structure that stores radiance field parameters for fast querying, while Plenoxels~\cite{plenoxels2022} and TensoRF~\cite{tensorf2022} replace neural networks with voxel grids or tensor factorization schemes to reduce computational overhead. 
Instant-NGP~\cite{mueller2022instant} introduces a multiresolution hash-grid positional encoding and integrates it into a CUDA/C++ library to accelerate the training and rendering of neural graphics primitives, establishing itself as a de facto standard for efficient, high-performance NeRFs.

Complementary research has investigated reducing memory requirements while preserving reconstruction fidelity.
BiRF~\cite{birf2023} and CNC~\cite{cncnerf2024} propose compact network architectures that trade memory footprint for additional computation.
While they better accommodate memory-constrained devices, the additional computation introduced by the more complex encoding strategies increases 
the execution time.
While instances of rendering on commodity GPUs can be found in literature, such as SNeRG~\cite{hedman2021snerg} and MobileNeRF~\cite{mobilenerf2023}, training requires access to powerful GPUs in existing methods, a significant barrier for NeRF-based approaches on embedded robotics devices. 
Table~\ref{tab:nerf-background} summarizes novel synthesis view methods, reporting their key contributions, average performance on the Synthetic NeRF 360° dataset \cite{mildenhall2020nerf}, measured with PSNR and structural similarity index measure (SSIM), parameter count, training time, and power consumption.

\begin{figure*}
    \centering
    \includegraphics[width=\linewidth]{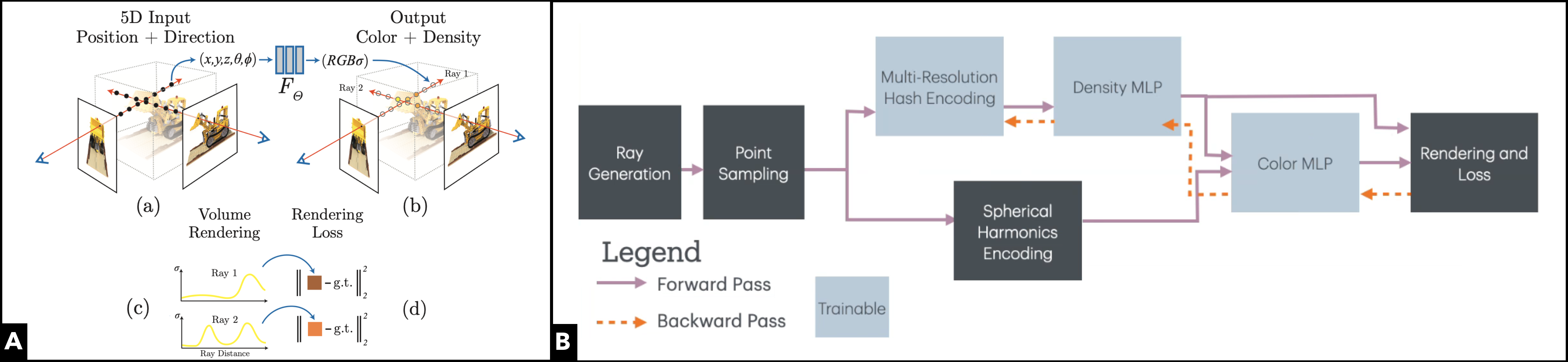}
    \caption{A) Novel view synthesis with neural radiance fields and volumetric rendering~\cite{mildenhall2020nerf}. B) The InstantNGP algorithm.}
    \label{fig:nerf-architecture}
\end{figure*}

\subsection{Federated learning for neural radiance fields}
Federated learning (FL)~\cite{mcmahan2017communication} is a collaborative machine learning framework in which multiple parties jointly train a shared model without sharing their raw data. Instead of exchanging data, participants send model updates to a coordinator, which may be either a dedicated server or a peer that also assumes the coordinating role. Training proceeds iteratively, alternating between local training phases, in which each participant updates the model on its private data, and communication phases, during which the coordinator aggregates the received updates into a global model and redistributes it to all participants to initialize the next round.

The first federated implementation of NeRF was proposed in FedNeRF~\cite{holden2023federated}, followed by studies investigating FL-based approaches for large-scale outdoor scene modeling~\cite{suzuki2023federated}, distributed multi-robot training with joint pose refinement~\cite{dinerf2024}, and planetary multi-agent mapping~\cite{szatmari2025federated}. These contributions remain exploratory, with no reported in-field deployment or detailed resource feasibility analysis.

Across these works, the main motivations for adopting FL have been scalability to larger datasets~\cite{suzuki2023federated,szatmari2025federated,tasneem2024decentnerfs}, maintainability of a dynamic process~\cite{suzuki2023federated}, and communication efficiency~\cite{dinerf2024,szatmari2025federated}. In contrast, in this work we employ FL primarily as a mechanism to enable a swarm of resource-constrained drones to collaboratively learn a model that outperforms what any single drone could achieve independently, thereby providing a stronger model for downstream tasks.

\section{Background}

\subsection{NeRF overview}
\renewcommand{\vec}[1]{\mathbf{#1}}
\newcommand{\mat}[1]{\mathbf{#1}}

A \textit{radiance field} represents a scene through an implicit function that maps spatial points $\vec{x}$ and view direction $\vec{d}$ to the volume density $\sigma$ and the emitted color $\vec{c}$:
\begin{equation}
    f_\theta(\vec{x}) = (\vec{c}, \sigma)
\end{equation}

\textit{NeRFs} approximate this function though two MLP networks, parametrized by $\theta = \{\tau, \phi\}$.
Instant-NGP~\cite{mueller2022instant}, shown in Figure~\ref{fig:nerf-architecture}, employs a \textit{density MLP} to predict the volume density $\sigma$ and a vector of hidden features $\vec{h}$ as a function of the 3D location $\vec{x}$:
\begin{equation}
    f_\tau(\gamma(\vec{x})) = (\vec{h}, \sigma),
\end{equation}
with 3D location represented with a position encoding $\gamma(\vec{x})$.

A \textit{color MLP} predicts the color $\vec{c}$ as a function of hidden features and viewing direction,
\begin{equation}
    f_\phi(\vec{h}, \gamma(\vec{d})) = \vec{c}.
\end{equation}
As with the 3D location, also the viewing direction is encoded as $\gamma(\vec{d})$.

The main contribution of Instant-NGP is the \textit{multi-resolution hash encoding}~\cite{mueller2022instant} (MRHE) to map the 3D position $\vec{x}$ into a high-dimensional parametric feature vectors $\gamma(\vec{x})$.
This trainable encoding is arranged into $L$ independent levels, each subdividing space in voxel grids of increasing resolution.
Each voxel is mapped to a hash table of up to $T$ feature vectors in $\mathbb{R}^B$.
A given location $\vec{x}$ is encoded to $\gamma(\vec{x}) \in \mathbb{R}^{L \times F}$, with trilinear interpolation from each layer.
As the resolution increases in deeper levels, multiple voxels map to the same feature vector in the fixed-size hash table but the subsequent density MLP learns to disambiguate the resulting aliasing. 
This key insight makes $T$ a scalable parameter we can scale to fit the tight memory requirements of our nano-drone.
Direction $\vec{d}$ is encoded as well, albeit with a coordinate-based (i.e., non trainable) encoding function $\gamma(\vec{d})$, the \textit{spherical harmonics}.

During training, \textit{samples}, composed of position and view directions, are generated by marching a set number of \textit{rays} $R$ through random pixels of random images from the train set and sampling points along the ray.
A \textit{batch} of $B$ samples, fed to the MLP, produces the sample colors and volume densities, which are aggregated using \textit{volume rendering} resulting in the color value of each ray, $\hat{\vec{c}}$.
Training optimizes the Huber loss of the ray color against the ground-truth pixel color using the Adam optimizer.
To reduce computation wasted on empty space, Instant-NGP fast forwards ray marching across empty regions through an
additional data structure, the \textit{occupancy grid}.
The occupancy grid is updated alternatively to network training, by querying density predicted by the density MLP in each cell of the reconstruction volume.

\subsection{Robotic platform}
\label{subsec:robotic_platform}

\begin{figure}[bt]
    \centering
    \includegraphics[width=\linewidth]{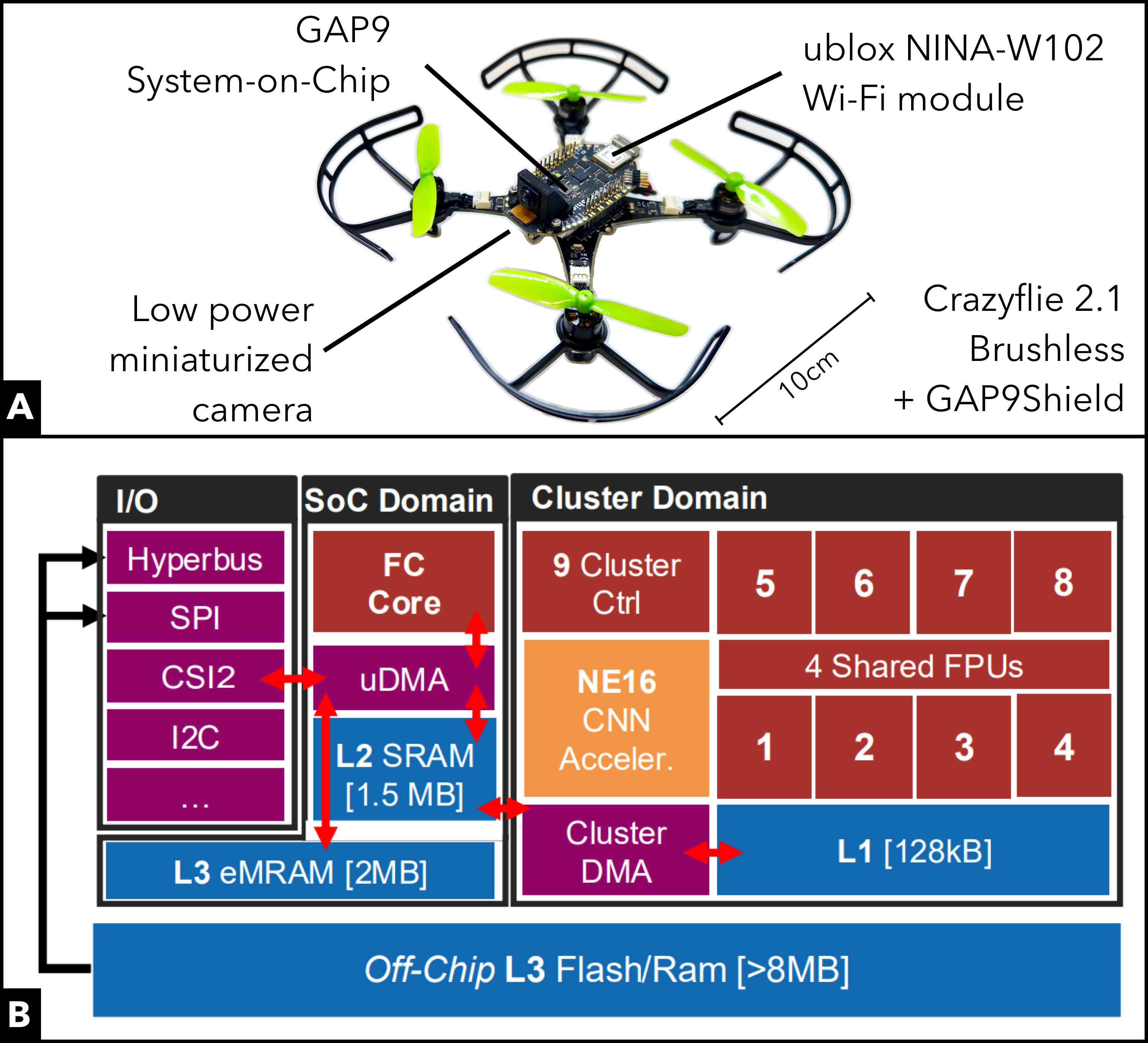}
    \caption{Target robot platform. A) Bitcraze Crazyflie 2.1 Brushless with GAP9Shield expansion deck. B) GWT GAP9 System-on-Chip architecture.}
    \label{fig:robotic_platform}
\end{figure}

Our robotic platform encompasses the Bitcraze Crazyflie 2.1 Brushless quadrotor (Figure~\ref{fig:robotic_platform}-A), a commercially available nano-drone that weighs \SI{27}{\gram} and measures \SI{10}{\centi\meter} in diameter. 
The drone is powered by a \SI{350}{\milli\ampere\hour} LiPo battery.
We extend this platform with the open-source \textit{GAP9Shield}~\cite{pcb_gap9shield}, a pluggable printed circuit board that integrates the GWT GAP9 System-on-Chip (SoC), \SI{32}{\mega\byte} of external L3 HyperRAM, a ublox NINA-W102 Wi-Fi transceiver (based on ESP32) providing up to \SI{15}{\mega\bit/\second} of bandwidth, and a low-power miniaturized camera.

The GAP9 SoC, depicted in Figure~\ref{fig:robotic_platform}-B, is organized into two power and frequency domains: the SoC domain, featuring a single main core called Fabric Controller (FC), and the Cluster, featuring 9 general-purpose cores, 4 floating-point units (FPUs) supporting 16- and 32-bit data types, and a mixed-precision AI hardware accelerator (NE16).
The Cluster cores support single-instruction multiple-data (SIMD) operations: the 9 general-purpose cores provide four-lane 8-bit integer SIMD, while the FPUs support two-lane 16-bit SIMD, in addition to scalar 32-bit floating-point operations.
The NE16 AI hardware accelerator provides a peak throughput of 150 integer \si{\giga\op/\second}.
However, it is designed for the forward pass of convolutional neural networks with integer data types, while our workload requires backpropagation and floating-point arithmetic.
In all experiments, we operate GAP9 at its peak performance configuration, clocked at \SI{370}{\mega\hertz} and powered at \SI{0.8}{\volt}, which translates to \SI{60}{\milli\watt} of average power consumption. 

The SoC features a hierarchical memory system with \SI{128}{\kilo\byte} of L1 memory shared across the Cluster and \SI{1.5}{\mega\byte} of L2 RAM accessible by both the FC and the Cluster.
The GAP9Shield further extends the system with \SI{32}{\mega\byte} of off-chip L3 OctaSPI RAM.
Cluster cores access L1 in around 1 cycle, while direct accesses to L2 incur an additional latency of about 100 cycles.
Two dedicated direct memory access (DMA) engines enable asynchronous data transfers, freeing the SoC cores to perform computation: the main DMA handles L3–L2 transfers (\SI{370}{\mega\byte/\second}), while the cluster DMA handles L2–L1 transfers (\SI{13.3}{\giga\byte/\second}).
By overlapping DMA transfers with computation, L2 data can be pre-fetched into L1 in the background, allowing the Cluster to operate exclusively on L1 and hide the L2 access latency.

\section{Methodology} \label{sec:methodology}

\subsection{Hardware-aware memory-efficient NeRF}
\label{subsec:methodology_shrinking}
As our robotic platform (Section~\ref{subsec:robotic_platform}) relies on a resource-constrained MCU with very limited on-chip and off-chip memory, the memory and compute demands of SoA novel view synthesis algorithms must be drastically reduced.
In its baseline configuration, Instant-NGP requires \SI{527}{\mega\byte} of memory and \SI{17.5}{\giga Ops} per training step.
The memory footprint, in particular, includes storage for both the model's learnable weights and the intermediate tensors needed for backpropagation. 
To shrink both peak memory footprint and operations per step, we perform a thorough hyperparameter optimization on Instant-NGP~\cite{mueller2022instant}, which targets
\textit{i}) the number of images per training step (img/step),
\textit{ii}) input image resolution,
\textit{iii}) batch size ($B$), and
\textit{iv}) maximum number of parameters per level of the MRHE ($T$).

We first analyze the number of images per training step (img/step). 
At each step, the baseline samples randomly from all 100 training images of the NeRF Synthetic dataset~\cite{mildenhall2020nerf}. 
On our MCU, storing 100 images would exceed on-chip capacity, forcing frequent off-chip accesses with much higher latency per access (see Section~\ref{subsec:robotic_platform}). 
Thus, minimizing the number of img/step (i.e., to 1 img/step) is desirable to reduce the memory transfer latency.
Next, we study the effect of input image resolution. 
Since our drone uses a low-resolution camera, we evaluate smaller inputs compared to the 800$\times$\SI{800}{\pixel} images of NeRF Synthetic. 
While Instant-NGP’s operations and peak memory do not directly depend on input resolution, larger images increase storage requirements because they must be stored in memory (typically off-chip) for the entire training process. 
Reducing the input size to 160$\times$\SI{160}{\pixel} lowers L3 memory utilization by \SI{61.44}{\mega\byte}.

We then ablate different batch sizes ($B$).
The peak memory footprint of Instant-NGP is dominated by the intermediate feature maps of the MLPs. 
Reducing $B$ linearly decreases both the number of operations per training step and the size of the output feature maps, significantly lowering peak memory requirements.
Last, we evaluate the effect of $T$. 
This parameter has negligible impact on the number of operations, since the number of hash queries per coordinate remains constant. 
However, it directly influences memory usage, as smaller $T$ values decrease the size of the hash tables.
Overall, the batch size $B$ is the primary factor controlling the number of operations per step, and thus total latency, while both $B$ and $T$ strongly affect Instant-NGP's total memory footprint.

\subsection{NeRF deployment}
\label{sec:methodology_deployment}

After optimizing Instant-NGP to reduce its operations and  memory footprint (a hard constraint of 32MB to deploy the model on our resource constrained MCU) through our shrinking methodology~\ref{subsec:methodology_shrinking}, we need an optimized embedded C implementation of the Tiny-DroNeRF algorithm.
The strict resource limitations of our platform require careful structuring of computation and memory transfers to fully exploit the computational capabilities of the GAP9.

Our embedded implementation is composed of
\textit{i}) handwritten kernels for each algorithmic block,
\textit{ii}) a tiling strategy, i.e., subdividing processed data into tiles that fit in local memory, and
\textit{iii}) batching, to unlock efficient parallel execution on the multi-core cluster.
We map each block of the Instant-NGP pipeline (Figure~\ref{fig:nerf-architecture}) to a custom hand-written kernel, which we parallelize across the GAP9 cluster. 
The baseline Instant-NGP uses \texttt{float16} arithmetic (except for the weight updates, accumulated in \texttt{float32} for stability~\cite{micikevicius2018mixedprecisiontraining}), which allows us to exploit GAP9’s optimized SIMD instructions for efficient computation of the MLP.

Although the configuration identified by the shrinking methodology fits within the total memory available on the nano-drone, we must ensure that each kernel’s inputs, outputs, and intermediate tensors reside in the \SI{128}{\kilo\byte} L1 memory. 
Otherwise, computation would stall on memory transfer bandwidth. 
To address this, we design a tiling strategy: subdividing kernel data into tiles that fit in L1 and scheduling memory transfers to overlap with computation, thereby keeping the cores consistently fed with data.

All kernels operate independently on batches of samples or rays, so memory accesses along these dimensions offer straightforward opportunities for parallelization and tiling. 
However, MRHE and occupancy grid would still not fit in the available L2; therefore, we leverage their hierarchical structure to move levels to L2 memory one at a time.

As the L3 memory bandwidth is about 35$\times$ lower than L2, minimizing L3 accesses is critical. 
Since most input data (rays and samples) is generated randomly, we restrict sampling to the maximum amount that can fit entirely within the on-chip L2 memory. 
However, training with excessively small batch sizes $B$ hinders convergence: it requires more steps and typically leads to lower reconstruction quality. 
To address this, we employ \textit{batch accumulation}, which decouples the size of computation batch tiles from the effective training batch size. 
In practice, we process batch tiles sized to fit in L2, while accumulating weight gradients over multiple steps until the desired effective batch size $B$ is reached.

\subsection{Federated implementation}

For the federated training of Tiny-DroNeRF, we employ a FL scheme in which one drone assumes a dual role, contributing to training with its own local data while simultaneously acting as the coordinator. The coordinator first performs a short pre-training phase of $N_{pt}$ steps on its local dataset and subsequently shares the resulting initial global model with the rest of the swarm.

During each communication round, all drones, including the coordinator, perform local training for $N_{\ell}$ steps and transmit their updated model parameters to the coordinator. We investigate two variants for parameter exchange: (i) transmitting the occupancy grid, hash table, and MLP parameters at every round, or (ii) transmitting only the hash table and MLP parameters, allowing each drone to update its occupancy grid locally.

Model aggregation is carried out using the FedAvg algorithm~\cite{mcmahan2017communication}, which computes a weighted average of the received parameters. In our case, since all drones possess an equal number of training images in our experiments, the weighted average simplifies to an arithmetic mean. The resulting global model is then broadcast to all drones, and the procedure is repeated over multiple communication rounds.

We evaluate federated training under two data partitioning setups. In the first setup, independent and identically distributed (IID), the training images are randomly shuffled and evenly partitioned across clients, so that each drone receives a representative sample of the entire dataset.
In the second setup, non-IID, we partition the training images by viewpoint sectors, so that each drone observes only a specific portion of the scene, leaving some sides completely unseen.
\section{Experimental results} \label{sec:results}

\subsection{Experimental setup}
We validate our models on the NeRF Synthetic 360° dataset~\cite{mildenhall2020nerf}, which contains 8 different objects, each one with 100 training images and 200 test images at a resolution of 800$\times$800.
To match the resolution of our drone cameras, we decrease the image resolution to 160$\times$160.
For evaluation, we follow prior work and report PSNR and SSIM~\cite{mildenhall2020nerf}.

\begin{figure}
    \centering
    \includegraphics[width=1\linewidth]{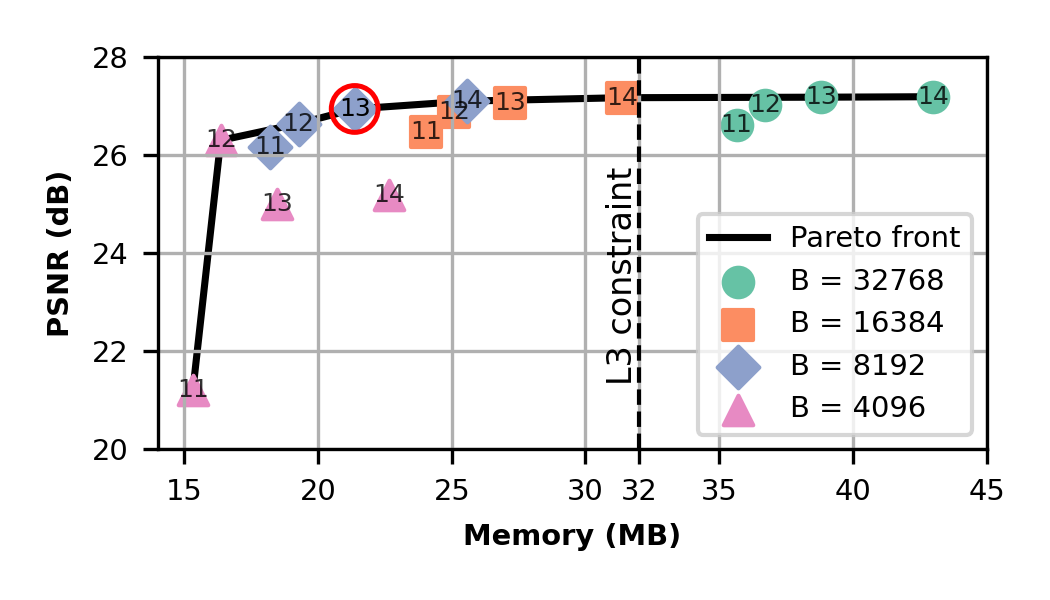}
    \caption{Memory vs. reconstruction quality Pareto curve.
    Each point corresponds to a configuration of batch size $B$ (marker) and encoding size $\log_2(T)$ (number).
    Circled in red, the configuration selected for deployment.
    }
    \label{fig:pareto}
\end{figure}

\subsection{Hardware-aware memory-efficient NeRF}\label{sec:hyper-search}

We assess how our shrinking methodology, through hyperparameter optimization, affects our system's key parameters, such as total memory footprint, operations per update step, and PSNR and SSIM reconstruction metrics.
Table~\ref{tab:hyperparam-search} explores various Instant-NGP's configurations, sweeping images per step, their resolution, the batch size ($B$), and the maximum number of parameters for a single level of the hash encoding ($T$).
For each table row, we keep the maximum PSNR and SSIM over five independent runs.
As expected, the memory required for the baseline Instant-NGP~\cite{mueller2022instant} (first row) would vastly exceed the total memory on our nano-drone, i.e., \SI{527}{\mega\byte} vs. \SI{32}{\mega\byte} at most.
Similarly, its training time would exceed two days, due to the \SI{17.52}{\giga Ops} per step.

A first optimization comes with the reduction in the number of images per step stored in the on-chip memory, passing from 100 to 1 image reduces memory from \SI{2.56}{\mega\byte} to \SI{0.03}{\mega\byte}, with a slight drop in PSNR of $3.6\%$.
To keep at least one level of the hash grid in our SoC's L2 memory, we need to reduce $T$ at least to $2^{14}$, despite all image resolution and batch size reductions.
Then, starting from this configuration (second row), which drops 23\% and 7\% in PSNR and SSIM, respectively, compared to the baseline, we perform a more fine-grained exploration of $B$ and $T$.

All configurations under investigation require the use of off-chip L3 memory. 
Therefore, we identify the configuration with $B=8\text{k}$ and $T = 2^{13}$ as the best trade-off, circled in Figure~\ref{fig:pareto}.
We name this configuration \textit{Tiny-DroNeRF}.

\begin{table}
\centering
\caption{Results of the hyperparameter optimization starting from the Instant-NGP model~\cite{mueller2022instant}. In bold, the configuration selected for deployment.}
\label{tab:hyperparam-search}

\setlength{\tabcolsep}{2.2pt}
\resizebox{\linewidth}{!}{
\begin{tabular}{
    c
    c
    c
    c
    c
    c
    c
    c
}
\toprule
{\multirow{2}{*}{\shortstack[c]{Img /\\ step}}} & 
{\multirow{2}{*}{\shortstack[c]{Res\\$[px]$}}} & 
{\multirow{2}{*}{$B$}} & 
{\multirow{2}{*}{$\log_2(T)$}} & 
{\multirow{2}{*}{\shortstack[c]{Mem\\$[\si{\mega\byte}]$}}} &
{\multirow{2}{*}{\shortstack[c]{Compute\\$[\si{\giga\op/step}]$}}} &
{\multirow{2}{*}{\shortstack[c]{PSNR\\$[\si{\deci\bel}]$}}} &
{\multirow{2}{*}{\shortstack[c]{SSIM \\ $[-]$}}} \\
& \\
\midrule
100 & 800 & 264k & 19 & 527 & 17.52 & 32.59 & 0.96 \\
\midrule
{\multirow{4}{*}{1}} & {\multirow{4}{*}{160}} & {\multirow{4}{*}{32k}} & 14 & 43 & 2.64 & 27.19 & 0.89 \\
& & & 13 & 39 & 2.62 & 27.18 & 0.90 \\
& & & 12 & 37 & 2.61 & 27.03 & 0.89 \\
& & & 11 & 36 & 2.61 & 26.61 & 0.89 \\ 
\midrule
{\multirow{4}{*}{1}} & {\multirow{4}{*}{160}} & {\multirow{4}{*}{16k}} & 14 & 31 & 1.65 & 27.17 & 0.89 \\
& & & 13 & 27 & 1.63 & 27.07 & 0.89 \\
& & & 12 & 25 & 1.62 & 26.87 & 0.89 \\
& & & 11 & 24 & 1.62 & 26.46 & 0.89 \\ 
\midrule
{\multirow{4}{*}{\textbf{1}}} & {\multirow{4}{*}{\textbf{160}}} & {\multirow{4}{*}{\textbf{8k}}} & 14 & 26 & 1.15 & 27.10 & 0.89 \\
& & & \textbf{13} & \textbf{21} & \textbf{1.14} & \textbf{26.94} & \textbf{0.89} \\
& & & 12 & 19 & 1.13 & 26.64 & 0.89 \\
& & & 11 & 18 & 1.11 & 26.17 & 0.88 \\ 
\midrule
{\multirow{4}{*}{1}} & {\multirow{4}{*}{160}} & {\multirow{4}{*}{4k}} & 14 & 23 & 0.91 & 25.18 & 0.78 \\
& & & 13 & 18 & 0.89 & 25.00 & 0.78 \\
& & & 12 & 16 & 0.88 & 26.30 & 0.89 \\
& & & 11 & 15 & 0.88 & 21.20 & 0.56 \\
 \bottomrule
\end{tabular}
}
\end{table}

\subsection{Deployment}

We analyze the execution time of Tiny-DroNeRF training on the GAP9 SoC.
We hand-write all computation kernels, designing parallelization and tiling strategies according to Section~\ref{sec:methodology_deployment} and profile their individual execution time on the cycle-accurate GAP9 GVSOC simulator.
We adopt $8\times$ \textit{batch accumulation}, processing 1024-sample batch tiles at a time and accumulating gradients up to the selected $B=\SI{8}{\kilo\nothing}$ effective batch size.
Peak on-chip memory usage across the entire pipeline remains within the tight GAP9 envelope.
The maximum allocations are \SI{1.3}{\mega\byte} in L2 and \SI{112}{\kilo\byte} in L1, fitting our on-chip memory limits (respectively \SI{1.5}{\mega\byte} and \SI{128}{\kilo\byte}).
Per-step L3 transfers amount to only \SI{3.6}{\mega\byte}, confirming excellent memory locality and ensuring that the transfer latency (\SI{9.7}{\milli\second}) does not dominate the step computation time (\SI{73}{\milli\second}).

\begin{figure}
    \centering
    \includegraphics[width=1\linewidth]{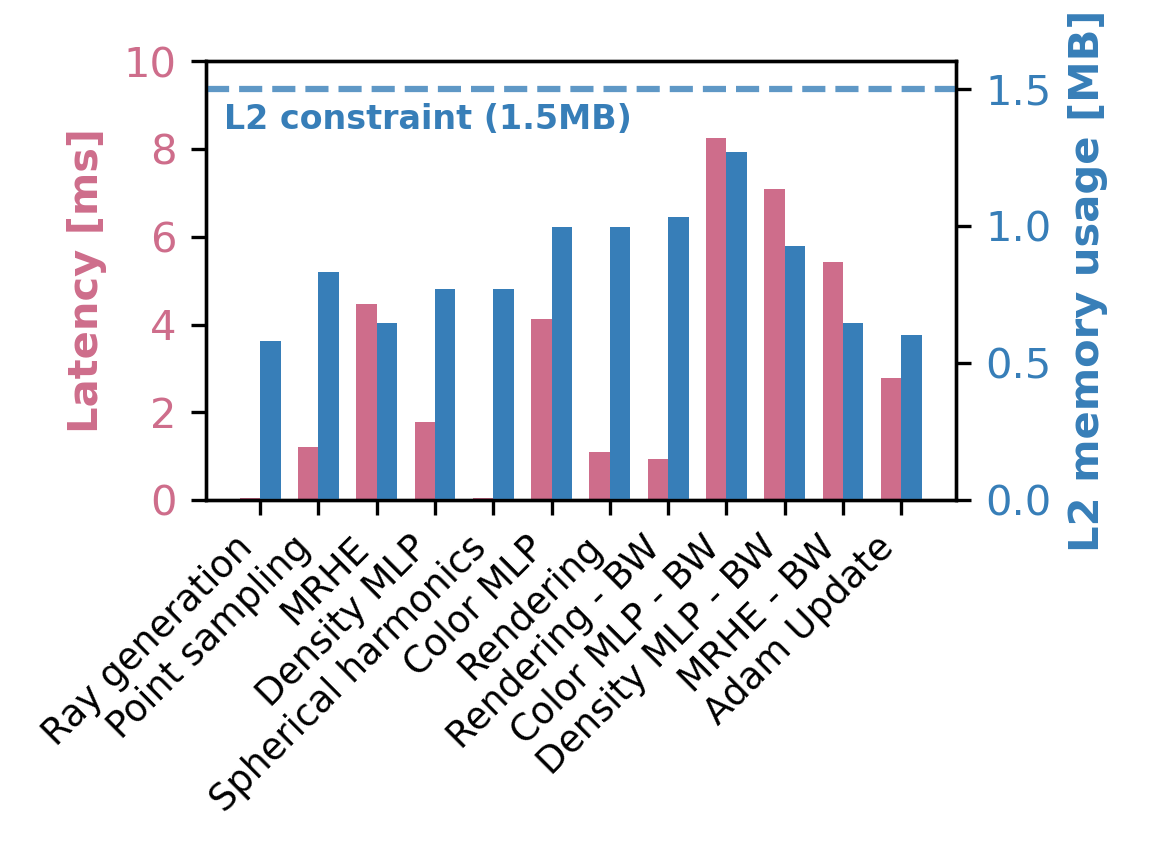}
    \caption{Kernel latencies and peak on-chip memory usage per training step over a 1024-sample batch tile.}
    \label{fig:latency_memory}
\end{figure}

Figure~\ref{fig:latency_memory} reports the execution latency (blue) and L2 memory usage (red) of each kernel on a single 1024-sample batch tile.
The latency amounts to \SI{12.8}{\milli\second} for the forward and \SI{21.7}{\milli\second} for the backward pass, while Adam optimizer updates take \SI{22.4}{\milli\second} every 8 step.
The backward pass of the color MLP dominates both latency and peak L2 usage, setting the global memory bound.
Other stages, such as MRHE and density MLP backward, contribute less but remain non-negligible.
We also perform occupancy grid updates~\cite{mueller2022instant} every 256 steps, with a latency of \SI{6.4}{\second} (not shown in Figure~\ref{fig:latency_memory}).

The amortized latency of a training step is thus \SI{73}{\milli\second}, which translates to \SI{97}{\minute} for a complete reconstruction on a single nano-drone ($8 \text{ batch tiles} \times \SI{10}{\kilo\nothing}$ training steps).
On the other hand, rendering novel views of the scene (i.e., inference) requires only forward passes.
The rendering latency scales with the scene-dependent samples per pixel~\cite{mueller2022instant}.
In our experiments on NeRF Synthetic 360º, rendering a $160\times\SI{160}{\pixel}$ image requires on average $\SI{13}{samples/pixel}$, corresponding to 325 forward passes over 1024-sample batch tiles. 
Given a forward pass latency of \SI{12.8}{\milli\second} per batch tile, the total rendering time amounts to approximately \SI{4.1}{\second}.

Overall, our profiling highlights the effect of memory-centric optimizations, such as tiling and batch accumulation, that keep latency computation-bound rather than data-movement-bound.

\begin{figure}[tb]
\centering
\includegraphics[width=\linewidth]{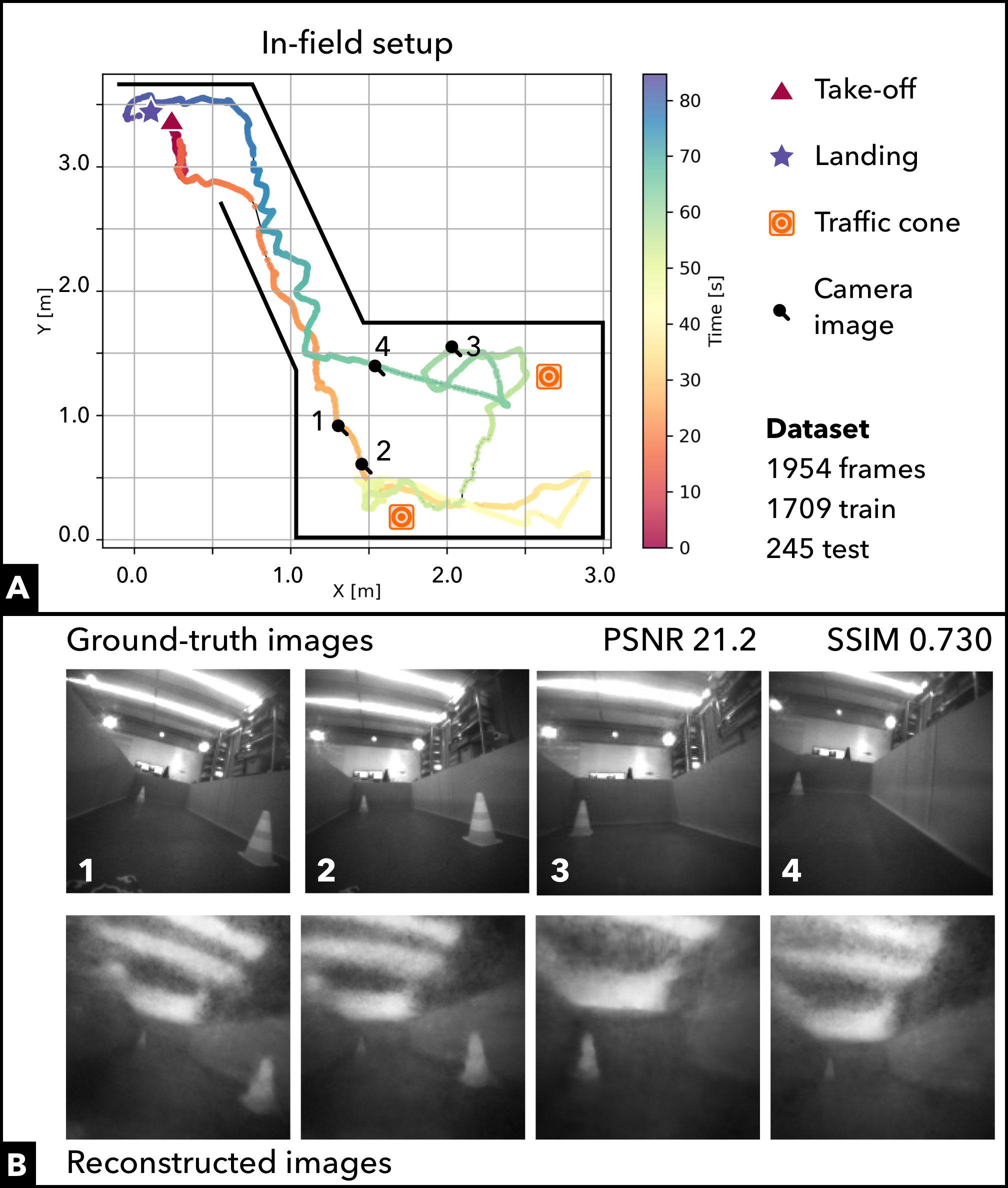}
\caption{A) In-field data collection setup. B) Samples of ground-truth images and the respective reconstructions.}
\label{fig:in-field}
\end{figure}

\subsection{In-field NeRF}

We evaluate Tiny-DroNeRF in a real deployment scenario by acquiring a new dataset with our nano-drone in an indoor environment featuring narrow walls and two traffic cones, as shown in Figure~\ref{fig:fed-results}-A. 
The drone flies across the space while recording video from its onboard camera. 
We acquire 160$\times$160\SI{}{\pixel} monochrome camera images at \SI{30}{\hertz}, streamed over Wi-Fi and post-processed with COLMAP~\cite{schoenberger2016sfm} to extract ground-truth camera poses. 
The resulting dataset contains 1954 frames, split into 1709 train and 245 test images.
Compared to the earlier experiments on NeRF Synthetic 360°, this in-field dataset introduces two additional challenges. 
First, images are grayscale rather than RGB, making reconstruction harder.
Second, the environment is cluttered and extends up to \SI{5}{\meter} in depth, unlike the synthetic dataset which features a single object with transparent background. 
This requires the model to reconstruct a larger scene volume. 

We train Tiny-DroNeRF on our in-field dataset and evaluate its reconstruction quality using PSNR and SSIM. 
Our model achieves a PSNR of 21.2 and an SSIM of 0.73. 
Figure~\ref{fig:in-field}-B illustrates qualitative results based on our supplementary video.
Ground-truth images are shown on the top, while reconstructions from Tiny-DroNeRF are shown on the bottom. 
We show that the reconstructed views capture the overall geometry of the corridor and the two cones in the environment, demonstrating that Tiny-DroNeRF can reconstruct real-world indoor scenes from grayscale images captured onboard our nano-drone.

\subsection{Federated learning}

\begin{figure}[tb]
    \centering
    \includegraphics[width=\linewidth]{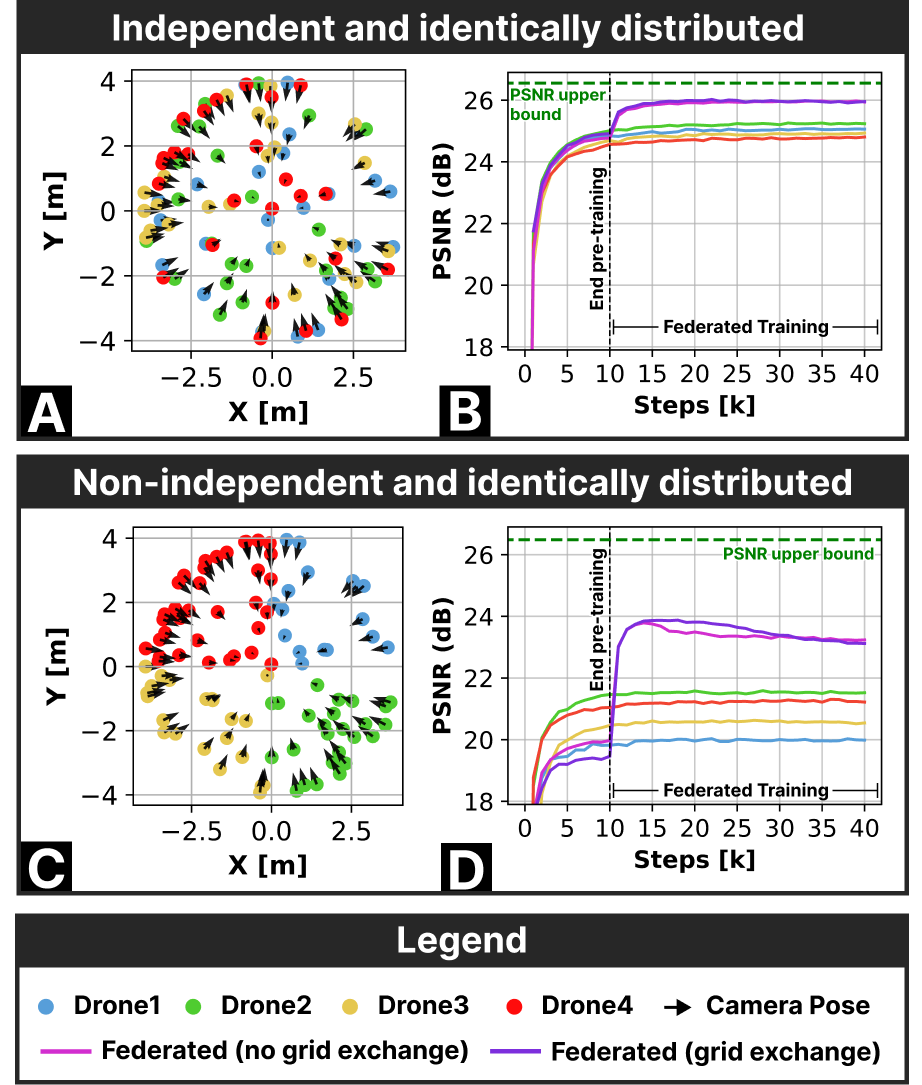}
    \caption{A-C) 2D top-view of the IID vs. non-IID dataset distributions, B-D) Federated learning PSNR testing curves for the IID vs. non-IID dataset trainings.}
    \label{fig:fed-results}
\end{figure}

We evaluated the federated implementation  of Tiny-DroNeRF in two distinct setups using the NeRF 360° Lego dataset~\cite{mildenhall2020nerf}. 
For both the IID and non-IID setups, we adhered to the dataset’s official train–test split, evenly distributing the 100 training images across four drones (25 per drone) and reserving the 200 test images exclusively for evaluation.
The corresponding data distributions are shown in Figure~\ref{fig:fed-results}-A-C.

Across both setups, we explored two pre-training configurations, $N_{pt}=5\,\mathrm{k}$ and $N_{pt}=10\,\mathrm{k}$ steps, to determine a pre-training duration that provides a sufficiently good initialization while avoiding delaying the start of federated training or causing unnecessary energy consumption. 
We also varied the number of local training steps per round, as excessively frequent communication prevents local models from making meaningful progress, whereas overly infrequent communication leads to local overfitting, with both ultimately slowing global convergence. Based on this exploration, we selected $N_{pt}=10\,\mathrm{k}$ for the pre-training steps and $N_{\ell}=1\,\mathrm{k}$ for the number of local training steps per round. All results in this section are based on this configuration.

We then compared models trained independently on each drone (on their own subset of images) with the two federated variants (with and without occupancy grid exchange) under both IID and non-IID conditions, as shown in Figure~\ref{fig:fed-results}-B-D.
For reference, the figure also includes a centralized model trained on the union of all local datasets, which represents an \textit{upper bound} on achievable performance.

Across both the IID and non-IID setups, the two federated variants, with or without occupancy grid exchange, achieve comparable results, indicating that exchanging the grid is unnecessary. 
This finding is particularly noteworthy, as the occupancy grid is a high-dimensional data structure and accounts for approximately 88\% of the total communication volume. Including the grid requires transmitting \SI{4.7}{\mega\byte} per round, resulting in a communication overhead of \SI{20.16}{\second} per round, whereas omitting it reduces the transmitted data to \SI{0.52}{\mega\byte} and the overhead to \SI{2.24}{\second} per round.

Furthermore, the federated models consistently outperform the individually trained models. 
In the IID setup, federated training achieves a PSNR that is \SI{0.7}{\decibel} higher than the best single-drone model and only \SI{0.6}{\decibel} below the centralized upper bound. 
In the more challenging non-IID setup, the federated models achieve performance \SI{3.25}{\decibel} below the upper bound but still \SI{1.7}{\decibel} above the best  single-drone model, demonstrating that through the federated model the drones can infer spatial characteristics of portions of the object they have not directly observed.

\section{Conclusion} \label{sec:conclusion}
Dense 3D reconstruction tasks have been unfeasible on nano-drones, as existing methods are not affordable with the available resources, i.e., sub-\SI{100}{\milli\watt} MCUs and a few tens of \SI{}{\mega\byte}.
We present Tiny-DroNeRF, a lightweight NeRF tailored for MCUs, which reduces memory by 96\% w.r.t to Instant-NGP with only a \SI{5.7}{\decibel} PSNR drop.
We reconstruct real-world scenes from onboard grayscale images at 21.2 PSNR. 
We further propose a federated learning scheme that allows swarms to synthesize unseen views from limited local data, improving PSNR by 1–\SI{2}{\decibel} over single-drone training with only \SI{2.24}{\second} of communication every \SI{1}{\kilo\nothing} steps.
To the best of our knowledge, this is the first demonstration of NeRF, with a federated learning approach, on a resource-constrained MCU.




\bibliographystyle{IEEEtran}
\bibliography{IEEEabrv,biblio}

\end{document}